\documentclass[10pt,twocolumn,letterpaper]{article}

\usepackage{cvpr}

\usepackage{booktabs}
\usepackage{multirow}
\usepackage{amsmath}
\usepackage{amssymb}
\usepackage{xcolor}
\usepackage{wrapfig}

\renewcommand{\Re}[0]{\mathbb{R}}

\usepackage{times}
\usepackage{epsfig}
\usepackage{graphicx}
\usepackage{moresize}
\usepackage{placeins}

\usepackage[pagebackref=true,breaklinks=true,letterpaper=true,colorlinks,bookmarks=false]{hyperref}
\usepackage{algorithmic}
\usepackage{algorithm}

\cvprfinalcopy 

\ifcvprfinal\fi
\begin{document}

\title{Supervised Mid-Level Features for Word Image Representation}

\author{Albert Gordo\\
Computer Vision Group\\
Xerox Research Centre Europe\\
{\tt\small albert.gordo@xrce.xerox.com}
}

\maketitle

\begin{abstract}
This paper addresses the problem of learning word image representations: given the cropped image of a word, we are interested in finding a descriptive, robust, and compact fixed-length representation.
Machine learning techniques can then be supplied with these representations to produce models useful for word retrieval or recognition tasks.
Although many works have focused on the machine learning aspect once a global representation has been produced, little work has been devoted to the construction of those base image representations:
most works use standard coding and aggregation techniques directly on top of standard computer vision features such as SIFT or HOG.

We propose to learn local mid-level features suitable for building word image representations.
These features are learnt by leveraging character bounding box annotations on a small set of training images. However, contrary to other approaches that use character bounding box information, our approach does not rely on detecting the individual characters explicitly at testing time.
Our local mid-level features can then be aggregated to produce a global word image signature.
When pairing these features with the recent word attributes framework of \cite{Almazan:2014}, we obtain results comparable with or better than the state-of-the-art on matching and recognition tasks using  global descriptors of only $96$ dimensions.
\end{abstract}
\section{Introduction}
\label{sec:intro}
In recent years there has been an increasing interest in tasks related to text understanding in natural scenes, and, amongst them, in word recognition: given a cropped image of a word, one is interested in obtaining its transcription.
The most popular approaches to address this task involve detecting and localizing individual characters in the word image and using that information to infer the contents of the word, using for example conditional random fields and language priors \cite{Wang:2010,Mishra:2012a,Mishra:2012b, Bissacco:2013, Neumann:2013}.
As shown by Bissacco \emph{et al.} \cite{Bissacco:2013}, such approaches that learn directly from the annotated individual characters can obtain impressive accuracy if large volumes of training data are available.
However, these approaches are not exempt from problems. 
First, to obtain a high accuracy, one needs to annotate very large amounts of words (in the order of millions) with character bounding boxes for training purposes, as done by Bissaco \emph{et al.} \cite{Bissacco:2013}. When limited training data is available, the recognition accuracy of these approaches is much lower \cite{Mishra:2012a, Mishra:2012b}.
Then, at testing time, one needs to localize the individual characters of the word image, which is slow and error prone.
Also, these approaches do not lead to a final signature of the word image that can be used for other tasks such as word image matching and retrieval.

Rather than localizing and classifying the individual characters in a word, a new trend in word image recognition and retrieval has been to describe word images with global representations using standard computer vision features (\eg HOG \cite{Dalal:2005}, or SIFT \cite{Lowe:2004} features aggregated with bags of words \cite{Csurka:2004} or Fisher vector \cite{Perronnin:2007} encodings) and apply different frameworks and machine learning techniques (such as using attribute representations, metric learning, or exemplar SVMs) on top of these global representations to learn models to perform tasks such as recognition, retrieval, or spotting \cite{Perronnin:2009,Rusinol:2011,Almazan:2012,Aldavert:2013,Almazan:2013,Rodriguez:2013,Almazan:2014}.
The global approaches have important advantages: they do not require that words be annotated with character bounding boxes for training and they do not require that the characters forming a word be explicitly localized at testing time.
They can also produce compact signatures which are faster to compute, store and index, or compare, while still obtaining very competitive results in many tasks.
The use of off-the-shelf computer vision features and machine learning techniques also makes them very attractive.
Yet, one may argue that not using character bounding box annotations during training, although very convenient, may be a limiting factor for their accuracy.

The main contribution of this paper is an approach to construct a global word image representation that unites the best properties of both main trends by leveraging character bounding boxes information at training time.
This is achieved by learning mid-level local features that are correlated with the characters in which they tend to appear.
We use a small external dataset annotated at the character level to learn how to transform small groups of locally aggregated low-level features into mid-level semantic features suitable for text, and then we encode and aggregate these mid-level features into a global representation. This unites advantages of both paradigms: a compact signature that does not require localizing characters explicitly at test time, but still exploits information about annotated characters.
Although some works have already used supervised information (in the form of text transcriptions) to project a global image representation into a more semantic space \cite{Rodriguez:2013, Almazan:2013},  
to the best of our knowledge, no other approach that constructs global image representations has leveraged character bounding box information at training time to do so.

We test our approach on two public benchmarks of scene text showing that constructing representations using mid-level features yields large improvements over constructing them using SIFT features directly.
When pairing these mid-level features with the recent attributes framework of \cite{Almazan:2013, Almazan:2014}, we significantly outperform the state-of-the-art in  word image retrieval (using both images or strings as queries), and obtain results comparable to or better than Google's PhotoOCR \cite{Bissacco:2013} in recognition tasks using a tiny fraction of training data: we use less than $5,000$ training words annotated at the character level, while \cite{Bissacco:2013} uses several millions.

The rest of the paper is organized as follows. Section \ref{sec:related} reviews the related work. Section \ref{sec:midlevel} describes our method. Section \ref{sec:experiments} deals with the experimental evaluation. Finally, Section \ref{sec:conclusions} concludes the paper.

\section{Related Work}
\label{sec:related}
We now review those works which are most related to our approach.
\paragraph{Scene text recognition.}
Most works focusing on scene-text target only the problem of recognition, \ie, given the image of a word, the goal is to produce its transcription. In such a case, a priori, there are no clear advantages with producing a global image representation. Instead, most methods aim at localizing and classifying characters or character regions inside the image and using this information to infer the transcription.
For example, Mishra \emph{et al.} \cite{Mishra:2012a, Mishra:2012b} propose to detect characters using a sliding window model and produce a transcription using a conditional random field model with language priors.
Neumann and Matas \cite{Neumann:2012, Neumann:2013} use Extreme Regions or Strokes to localize and describe characters. Words are then recognized using a commercial OCR or by maximizing the characters' probability.
The recent \cite{Yao:2014} uses a mid-level representation of strokes to produce more semantic descriptions of characters, that are then classified using random forests.
PhotoOCR \cite{Bissacco:2013} learns a character classifier using a deep architecture with millions of annotated training characters, and produces very accurate transcriptions at the cost of requiring vast amounts of annotated data.
In a different line, Jaderberg \emph{et al.} \cite{Jaderberg:2014b} propose to learn the classification task directly from the image without localizing the characters using deep convolutional neural networks. Although the results on some tasks are impressive, this approach also requires millions of annotated training samples to perform well.
\paragraph{Global representations for word images.} 
More directly related to our work are global representations. Producing  image signatures opens the door to other tasks such as word retrieval, as well as easing tasks such as storing and indexing word images.
Rusi\~nol \emph{et al.} \cite{Rusinol:2011} construct a bag of words over SIFT descriptors to encode word images, and use it to perform segmentation-free spotting on handwritten documents. In \cite{Aldavert:2013}, this framework is enriched using textual information. In \cite{Almazan:2012}, Exemplar SVMs and HOG descriptors are also used to perform segmentation-free spotting on documents. Goel \emph{et al.} \cite{Goel:2013} propose to recognize scene-text images by synthesizing a dataset of annotated images and finding the nearest neighbor in that dataset. Rodriguez and Perronnin propose in \cite{Rodriguez:2013} a label embedding approach that puts word images (represented with Fisher vectors) and text strings in the same vectorial space. Similarly, Almaz\'an \emph{et al.} \cite{Almazan:2013,Almazan:2014} propose a word attributes framework that can perform both retrieval and recognition in a low dimensional space. 
Although they perform well in retrieval tasks, they are outperformed in recognition by methods that exploit annotated character information, such as PhotoOCR \cite{Bissacco:2013}.

\label{sec:midlevel}
\begin{figure*}[t!]
\centering
\includegraphics[width=0.95\textwidth]{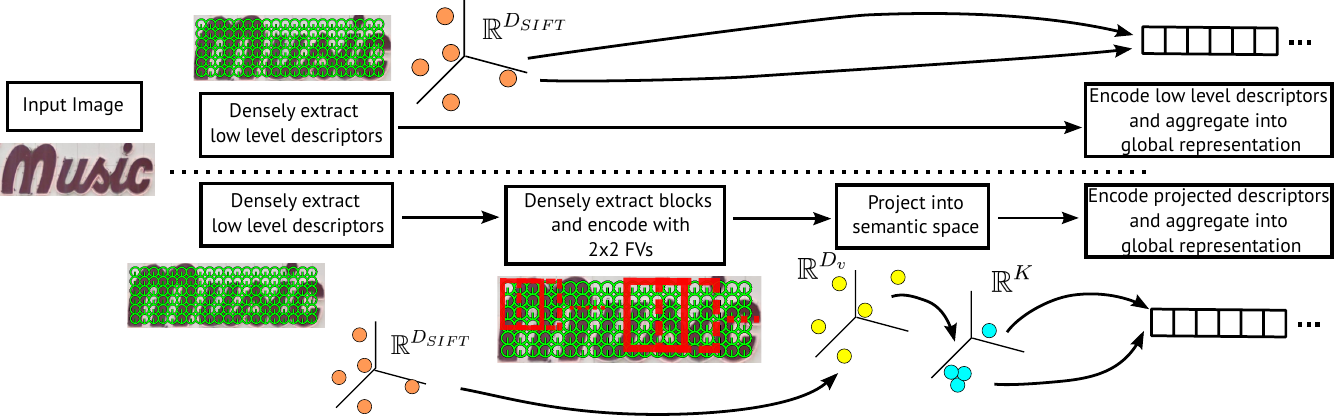}
\caption{\textbf{Top.} Standard word image description flow: low-level descriptors (\eg SIFT) are first densely extracted and then encoded and aggregated into a global representation using \eg Fisher vectors (FV). Spatial pyramids may be used to add some weak geometry. \textbf{Bottom.} Proposed approach: we first densely extract low-level descriptors.  Then we densely extract blocks of different sizes, and represent each block by aggregating the low-level descriptors it contains into a local FV with a $2\times 2$ spatial pyramid. These local FV representations are then projected into a mid-level space correlated with characters. Finally, these mid-level features are aggregated into a global FV.}
\label{fig:flow1}
\end{figure*}

\paragraph{Learning mid-level features.}
Our work is also related to the use of mid-level features (\eg \cite{Boureau:2010, Juneja:2013, Yao:2014}), where ``blocks'' that contain  some basic semantic information are discovered, learned, and/or  defined. The use of mid-level features has been shown to produce large improvements in different tasks. Of those works, the most related to ours is the work of Yao \emph{et al.} \cite{Yao:2014}, which learns Strokelets, a mid-level representation that can be understood as ``parts'' of characters. These are then used to represent characters in a more semantic way. The main distinctions between the Strokelets and our work are that i), we exploit supervised information to learn a more semantic representation, and ii), we do not explicitly classify the character blocks, and instead use this semantic representation to construct a high-level word image signature. We show that our approach leads to significantly better results than the Strokelets \cite{Yao:2014}.
The embedding approaches of \cite{Rodriguez:2013, Almazan:2013} could also be understood as producing supervised mid-level features, but do not use character bounding box information to do so.
To learn the semantic space, we perform supervised dimensionality reduction of local Fisher vectors that are then encoded and aggregated into a global Fisher vector.
This could be seen as a deep Fisher network for image recognition \cite{Simonyan:2013}, also used very recently for action recognition \cite{Peng:2014}.
The main difference is that in \cite{Simonyan:2013,Peng:2014} the supervised dimensionality reduction step is learned using the image labels, the same labels that will be used for the final classification step.
In our case, the goal is to transfer knowledge from the individual character bounding boxes, which are only annotated in the training set, to produce features that are correlated with characters, and exploit this information in the target datasets, where these bounding boxes are not available. This would be similar to learning the extra layer of the deep Fisher network using the labels of the bounding boxes of the objects, instead of using the whole image label as \cite{Simonyan:2013} does. Although the resulting architectures are similar, the motivation behind them is very different.
In that sense, our work can also be related to works on learning with privileged information \cite{Vapnik:2009, Sharmanska:2013, Gong:2013}, where the information available at training time to construct the representations (in our case, character bounding boxes) is not available at test time.

\section{Mid-Level Features for Word Images}
\label{sec:midlevel}
A standard approach to construct a global word image representation is to i) extract low-level descriptors, ii) encode the descriptors, and iii) aggregate them into a global representation, potentially with spatial pyramids \cite{Lazebnik:2006} to add some weak geometry. This representation can then be used as input for different learning approaches, as done \eg in \cite{Rusinol:2011, Almazan:2013,Rodriguez:2013}. Figure \ref{fig:flow1} (top) illustrates this process.

In our proposed method, we aim at constructing global representations based on semantic mid-level features instead of using the low-level descriptors directly. The goal is to produce features that might not be good enough to predict the individual characters directly, but are  more correlated with the individual characters than SIFT or other local descriptors.

This is achieved not by finding and classifying characteristic blocks as in \cite{Yao:2014}, but by projecting \emph{all} possible image blocks into a lower-dimensional space where our mid-level visual features and the characters are more correlated. By projecting all possible blocks in an image, one obtains a set of mid-level descriptors that can then be aggregated into a global image representation. The process is illustrated in Figure \ref{fig:flow1} (bottom).

In what follows, we first describe the learning process and describe how to project the image blocks into the semantic space correlated with the characters (Section \ref{sec:learning}).
We then describe how to extract these mid-level features from a new image, and how to combine them with the word attributes framework \cite{Almazan:2014} to obtain very compact, discriminative word representations (Section \ref{sec:representation}).

\subsection{Learning Mid-Level Features}
\label{sec:learning}
Let us assume that, at training time, one has access to a set of $N$ word images and their respective annotations.
Let $\mathcal{I}$ be one word image containing $|\mathcal{I}|$ characters, and let its annotation be a list of character bounding boxes $char\_bb$ and character labels $char\_y$, $\mathcal{C}_\mathcal{I} = \{(char\_bb_i,char\_y_i), i=1\ldots |\mathcal{I}|\}$. Each character label $char\_y_i$ belongs to one of the $62$ characters in the following alphabet: $\Sigma = \{\verb|A|,\ldots,\verb|Z|,\verb|a|,\ldots,\verb|z|,\verb|0|,\ldots,\verb|9|\}$. 

Let us denote with $block\_bb$ a square block in the image represented as a bounding box. For training purposes, let us randomly sample from every image $S$ blocks of different sizes (\eg $32 \times 32$ pixels, $48\times 48$, etc) at different positions.
Some of these blocks may contain only background, but most of them will contain parts of a character or parts of two consecutive characters.

We then describe these blocks using two modalities. The first one is based on visual features. The second one is based on character annotations. This second modality is more discriminative but is only available during training.

Then, one can learn a mapping between the visual features and the character annotation space. Once this mapping has been learned, at testing time one can extract all possible blocks in an image, represent them first with low-level visual features, and then map them into this semantic space to obtain mid-level features, as seen in Figure \ref{fig:flow1} bottom.

\paragraph{Block visual features.} To encode the visual aspect of the blocks we use Fisher vectors (FV) \cite{Perronnin:2007} over SIFT descriptors \cite{Lowe:2004}, with a $2\times 2$ spatial pyramid \cite{Lazebnik:2006} to add some basic structure to the block. The descriptors are then $\ell_2$-normalized.
It has been shown that power- and $\ell_2$- normalizing the Fisher vector usually leads to more discriminative representations \cite{Perronnin:2010}. In this case, however, we only apply an $\ell_2$ normalization: the non-linearity introduced by the power-normalization would make it more difficult to efficiently aggregate the block statistics (see Section \ref{sec:representation} and appendix for details).
Fortunately, power normalization is most useful when using large vocabularies \cite{Perronnin:2010b}. With small vocabularies ($8$ Gaussians in our case) the improvements due to power normalization are very limited.

Finally, stacking the descriptors of all the sampled blocks of all the training images leads to a matrix $\mathbf{X}$ of size  $NS \times D_v$, where we denote with $D_{v}$ the dimensionality of these visual representations. In our experiments we will use visual descriptors of $D_{v}=4,096$ dimensions. 

\begin{figure}[t!]
\centering
\includegraphics[width=0.8\linewidth]{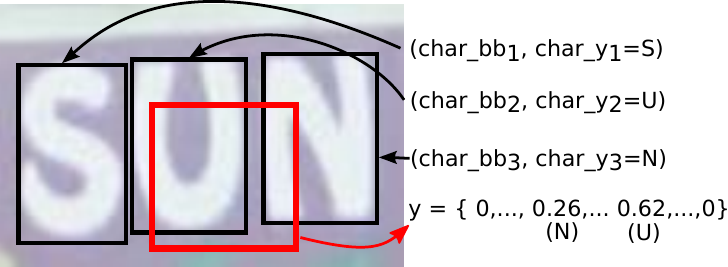}
\caption{
Example of annotated word and a sampled block with its label. The characters of the word contain bounding boxes (in black) and label annotations ('S','U','N'). The image also shows a sampled block (in red) with its respective computed label $y$. All the elements of $y$ are set to $0$ except the ones corresponding to the 'U' and 'N' characters.}
\label{fig:annotation}
\end{figure}

\paragraph{Block annotations.} The second view is based on the annotation and contains information about the overlaps between the blocks and the characters in the word image.
The goal is to encode with which characters the sampled blocks tend to overlap. In particular, we are interested in encoding what percentage of the character regions are covered by the blocks.
As a first approach, we construct a $D_a$-dimensional label $y$ for each sampled block, with $D_a = | \Sigma | = 62$.
Given a block, its label $y$ is encoded as follows: for each character $\Sigma_d$ in the alphabet $\Sigma$, we assign at dimension $d$ of the label vector the normalized overlap between the bounding box of the block and the bounding boxes of the characters of the word whose label $char\_y_i$ equals $\Sigma_d$. Since the word may have repeated characters overlapping with the same block, we take the maximum overlap:
\begin{equation}
\small
y^d = \max_{(char\_bb_i,char\_y_i) \in \mathcal{C}_\mathcal{I}} \delta_{i,d} \frac{|\text{Intersection}(block\_bb, char\_bb_i)|}{|char\_bb_i|},
\label{eq:labels}
\end{equation}
where $|\cdot|$ represents the area of the region, and $\delta_{i,d}$ equals $1$ if $char\_y_i=\Sigma_d$ and $0$ otherwise.
Figure \ref{fig:annotation} illustrates this with an annotated image and a sampled block with its corresponding computed label $y$.

As with the first view, it is possible to encode all the block labels of all the sampled blocks of all the training images in a matrix $\mathbf{Y} \in \Re^{NS \times D_a}$.

\textbf{Adding character spatial information.} The labels that we introduced present a shortcoming: although they correlate blocks with characters, they do not encode \emph{which part of the character} they are correlated with. This is important, since this type of information can have great discriminative power. 
To address this problem, we split the ground-truth bounding box character annotations in $R\times R$ character regions (CR). One can consider a $1\times 1$ region (\ie, the whole character), but also $2\times 2$ regions, $3\times 3$ regions, etc.
Then, the labels $y$  encode the overlap of the block with each region of the character independently, leading to a label of size $\Re^{R^2\times D_a}$:
\begin{equation}
\small
y^{d,r} = \max_{(char\_bb_i,char\_y_i) \in \mathcal{C}_\mathcal{I}} \delta_{i,d} \frac{|\text{Intersection}(block\_bb, char\_bb_{ir})|}{|char\_bb_{ir}|},
\label{eq:labels2}
\end{equation}
with $r=1\ldots R^2$, and where $char\_bb_{ir}$ denotes the $r$-th region of bounding box $char\_bb_{i}$. This label is flattened into $\Re^{R^2D_a}$ dimensions. One may also consider computing labels at different character levels and concatenating the results in a final label before learning the projection, but we did not find this to improve the results.

\paragraph{Learning the mid-level space.} To find a mapping between $\mathbf{X}$ and $\mathbf{Y}$ one can use, for example,  canonical correlation analysis (CCA) \cite{Hotelling:1936}.
CCA finds two projection matrices $\mathbf{U} \in \Re^{D_v\times K}$ and $\mathbf{V} \in \Re^{D_a \times K}$ such that the correlation in the projected space is maximized. To find $\mathbf{U}$, one only needs to solve a generalized eigenvalue problem. To avoid numerical instabilities, a small regularization parameter $\eta$ is typically used. When $D_v$ and $D_a$ are small as in our case, this is very fast, and needs to be solved only once, offline. The $K$ leading eigenvectors of the solution constitute the columns of matrix $\mathbf{U}$. One can easily choose the output dimensionality $K$ by keeping only a certain number of eigenvectors.
Analogously, one can solve a related eigenvalue problem to find $\mathbf{V}$, although we do not use it here since we will not have access to character bounding boxes at test time.
This process allows one to learn a matrix $\mathbf{U}$ that projects $\mathbf{X}$ into a subspace of $K$ dimensions that is correlated with the labels $\mathbf{Y}$. This is depicted in Algorithm \ref{alg:alg1}.

\begin{algorithm}[t!] 
\caption{Learn embedding} 
\begin{algorithmic} 
\footnotesize
\REQUIRE Training images $\boldsymbol{\mathcal{I}}= \{\mathcal{I}_1, \ldots, \mathcal{I}_N\}$ and their annotations.
Output dimensionality $K$. Number of sampled blocks per image $S$.
Regularization parameter $\eta$.
\ENSURE  Embedding matrix $\mathbf{U}$.
\FOR{$\mathcal{I} \in \boldsymbol{\mathcal{I}}$}
\STATE Densely extract SIFT descriptors.
\STATE Sample $S$ blocks at different sizes and positions.
\FOR{each block}
\STATE \textbf{i)} Aggregate the SIFT descriptors inside the block bounding box into a FV with a $2\times 2$ spatial pyramid. Add as a new row of $\mathbf{X}$.
\STATE \textbf{ii)} Construct label of block using Equation \eqref{eq:labels} or Equation \eqref{eq:labels2}. Add as a new row of $\mathbf{Y}$.
\ENDFOR
\ENDFOR
\STATE Compute $\mathbf{U}$ using CCA. The $k$-th column of $\mathbf{U}$ is the eigenvector solution of the generalized eigenvalue problem $\mathbf{X}^T\mathbf{Y}(\mathbf{Y}^T\mathbf{Y + \eta I})^{-1}\mathbf{Y}^T\mathbf{X}   u_k =  \lambda(\mathbf{X}^T\mathbf{X} + \eta I)u_k$ associated with the $k$-th largest eigenvalue.
\STATE \hspace{-0.4cm} \textbf{End}
\end{algorithmic}
\label{alg:alg1} 
\end{algorithm}

\paragraph{Discussion.} In the described system, the block labels encode which percentage of the characters or the character regions are covered by the blocks. However, this is only one possible way to encode the labels. Other options could include \eg encoding which percentage of the block is covered, the intersection over union, or working at the pixel level instead of the region level. Any representation that relates the visual block with the character annotation could be considered.  We found that the proposed approach worked well in practice and deemed the search for the optimum representation out of the scope of this work.

\subsection{Representing Word Images with Mid-Level Features}
\label{sec:representation}
Once the matrix $\mathbf{U}$ has been learned, one can use it to compute the set of mid-level features of a new word image.
A naive approach would involve extracting all possible blocks, encoding them with Fisher vectors, and projecting them with $\mathbf{U}$ into the mid-level space. This is shown in Algorithm \ref{alg:alg2}.

\textbf{Fast computation.} In practice, however, applying this algorithm directly would be slow, requiring one to encode all possible blocks independently.
Instead, we propose an approach to compute these descriptors \emph{exactly} in an efficient manner.
The key aspect is to notice that if the block Fisher vectors are not power-normalized (as in our case), all operations involved in the computation of the mid-level features are linear: one can compute the mid-level representation of each individual SIFT descriptor, and the mid-level representation of a block is exactly the $\ell_2$-normalized sum of the mid-level representations of the SIFT descriptors contained in that block.
Therefore, one can i) extract all SIFT descriptors of the image, ii) compute their mid-level representation, and iii) aggregate into an integral representation.
Given that structure, one can compute the mid-level descriptor of an arbitrary block exactly with just two sums and two subtractions of low-dimensional vectors plus a final $\ell_2$ normalization.
A more detailed description of the efficient aggregation approach is provided in the appendix.

\textbf{Building word representations.} The mid-level features can then be encoded and aggregated into a global image representation using \eg Fisher vectors.
These global image representations can then be used by themselves, but can also be used as building blocks to more advanced methods that use global image signatures as input (\eg, \cite{Rodriguez:2013, Almazan:2014}).
We focus on the recent \emph{word attributes} work of Almaz\'{a}n \emph{et al.} \cite{Almazan:2014}, with available source code and state-of-the-art results in word image matching.
This work uses Fisher vectors on SIFT descriptors as a building block to predict character attributes. The character attributes represent the presence or absence of a given character at a given relative position of the word (\eg, ``word contains an \emph{a} in the second half of the word'' or ``word contains a \emph{d} in the first third of the word'').
Word images are then described by the predicted attribute scores.
This attribute representation is then projected into an embedded space correlated with embedded text strings using CCA, which improves its discriminative power while reducing its dimensionality. In our experiments, we will use this to produce global image signatures of only $96$ dimensions.

A great advantage of this framework is that representations of images and strings can then be compared using a cosine similarity, providing a unified framework to perform query-by-example (QBE) matching (\ie, retrieve images of a dataset given a query image), query-by-string (QBS) matching (\ie, retrieve images of a dataset given a query text string), and recognition (\ie, retrieve text strings given a query image).
Since the approach is already based on Fisher vectors, it is easy to replace the SIFT descriptors in the pipeline of \cite{Almazan:2014} with our mid-level features and measure exactly their contribution.

\begin{algorithm}[t!] 
\caption{Extract mid-level features} 
\begin{algorithmic} 
\footnotesize
\REQUIRE Input image  $\mathcal{I}$.
Embedding matrix $\mathbf{U}$.
Block sizes (\eg $16\times 16$, $32 \times 32$, etc).
Block step size (\eg 4 pixels).
\ENSURE  Mid-level features of image $\mathcal{I}$
\FOR{\textbf{each block size}}
\FOR{\textbf{each possible block position inside the image (according to the block step size)}}
\STATE \textbf{i)} Densely extract SIFT descriptors inside the block.
\STATE \textbf{ii)} Encode and aggregate the SIFT descriptors into a FV with a $2\times 2$ spatial pyramid and $\ell_2$ normalize.
\STATE \textbf{iii)} Project the FV with $\mathbf{U}$ and $\ell_2$ normalize again to construct the mid-level feature of the block.
\ENDFOR
\ENDFOR
\STATE Return all mid-level features along with the position and scale of the blocks. 
\STATE \hspace{-0.4cm} \textbf{End}
\end{algorithmic}
\label{alg:alg2} 
\end{algorithm}

\section{Experiments}
\label{sec:experiments}
We start by describing the data used for learning the mid-level features and for evaluation purposes.
We then describe the evaluation protocols and report the experimental results.

\subsection{Datasets}\label{sec:datasets}
\paragraph{Evaluation datasets.} We evaluate our approach on two public benchmarks: IIIT5K \cite{Mishra:2012a} and Street-View Text (SVT) \cite{Wang:2011}. IIIT5K is the largest public annotated scene-text dataset to date, with $5,000$ cropped word images: $2,000$ training words and $3,000$ testing words. Each testing word is associated with a small text lexicon (SL) of $50$ words and a medium text lexicon (ML) of $1,000$ words used for recognition tasks. Note that each word has a different lexicon associated with it.
SVT is another popular dataset with about 350 images harvested from Google Street View. These images contain annotations of  $904$ cropped words: $257$ for training purposes and $647$ for testing purposes. Each testing image has an associated lexicon of $50$ words. 
We also construct a \emph{combined} lexicon (CL), that contains every possible word that appears in the lexicons of each dataset ($1,787$ unique words on IIIT5K and $4,282$ on SVT).

\paragraph{Learning dataset.}
Learning the proposed transformations requires gathering training words annotated at the character level.
Fortunately, pixel level annotations and character bounding boxes exist for several standard datasets \cite{Kumar:2012}. We gathered annotations for ICDAR 2003 \cite{Lucas:2003}, Sign Recognition 2009 \cite{Weinman:2009}, ICDAR 2011 \cite{Shahab:2011}, and IIIT5K \cite{Mishra:2012a} -- for IIIT5K, only the training set annotations were used.
In total, we gathered $3,829$ words annotated with approximately $22,500$ character bounding boxes that are used to learn the mid-level features transformation.



\subsection{Implementation details}
To construct our block FVs, we extract SIFTs \cite{Lowe:2004} at 6 different scales, project them with PCA down to $64$ dimensions, and aggregate them using FVs (gradients \emph{w.r.t.} means and variances)  with $8$ Gaussians and a spatial pyramid of $2\times 2$, leading to a dimensionality $D_v$ of $2\times 64 \times 8 \times 4 = 4,096$.
During training, we sample $150$ blocks per training image. In total, we sampled approximately $600,000$ blocks. To learn the projection matrix $U$ with CCA, we use a regularization of $\eta = 1e^{-4}$ in all our experiments.

To construct a global representation, we first extract all mid-level blocks at 5 block sizes ($16\times 16$, $24\times 24$, \ldots, $48 \times 48$) with a step size $p$ of $4$ pixels and $K=62$ CCA dimensions using the efficient approach described in Section \ref{sec:representation}.
Given images of $120$ pixels in height, on average, we can extract and describe all blocks in less than a second using a MATLAB implementation and a single core.
Then we append the normalized $x$ and $y$ coordinates of the center of the block as suggested by S\'anchez \emph{et al.} \cite{Sanchez2012}, and aggregate using a global FV with a $2\times 6$ spatial pyramid with  $16$ Gaussians, leading to $24,576$ dimensions. These global Fisher vectors are then power- and $\ell_2$- normalized \cite{Perronnin:2010}, and can be compared using the dot-product as a similarity measure. 

To construct the baseline global representation based only on SIFT (with no mid-level features) we follow a very similar approach, but instead of computing the mid-level blocks and reducing their dimensionality down to $62$ dimensions with CCA, we use the SIFT features directly, reducing their dimensionality down to $62$ dimensions with PCA and appending the normalized $x$ and $y$ coordinates of the center of the patch. This mimics the setup of the word attributes framework of \cite{Almazan:2014}.
We also experiment with reducing the dimensionality of our mid-level features in an unsupervised manner with PCA instead of CCA, to separate the influence of the extra layer of the architecture from the supervised learning.

When using word attributes, one has control of the final dimensionality of the representations. We set this output dimensionality to $96$ dimensions. We observed that, in general, accuracy reached a plateau around that point: after that, increasing the number of dimensions does not significantly affect the performance.
We found only one exception, where increasing the number of output dimensions beyond $96$ in one of the SVT experiments significantly improved the results.

\subsection{Evaluation}\label{sec:evaluation}
We evaluate our approach with two different setups.
In the first one, we are interested in observing the effect of using supervised and unsupervised mid-level features instead of SIFT descriptors directly when computing global word image representations, \emph{without applying any further supervised learning}. We compute Fisher vectors using a) SIFT features, b) unsupervised mid-level features (\ie, dimensionality reduction of the block FVs with PCA), and c) supervised mid-level features (\ie, dimensionality reduction with CCA). In this last case, we also explore the effect of the number of character regions (CR) used to compute the labels during training (\cf Equation \eqref{eq:labels2}),  from a $1\times 1$ to a $4\times 4$ region split. These FVs can be compared using the dot-product as a similarity measure.

We measure the accuracy in a query-by-example retrieval framework, where one uses a word image as a query, and the goal is to retrieve all the images of the same word in the dataset. We use each image of the test set in a leave-one-out fashion to retrieve all the other images in the test set. Images that do not have any relevant item in the dataset are not considered as queries. 

We report results in Table \ref{tab:baseline} using both mean average precision and precision at one as metrics. We highlight three aspects of the results. First, \emph{using supervised mid-level features significantly improves over using SIFT features directly}, showing that the representation encodes more semantic information. Second, \emph{the improvement in the mid-level features comes from the supervised information and not from the extra layer of the architecture}: the unsupervised mid-level features perform worse than the SIFT baseline. And, third, \emph{encoding \emph{which part} of the characters the blocks overlap with is more informative than only encoding which characters they overlap with}. We will use the $4\times 4$ split for the rest of our experiments.

\begin{table}[t!]
\footnotesize
\centering
\caption{Baseline results on query-by-example using Fisher vectors (FV). In the case of the supervised mid-level features, we evaluate the effect of the number of character regions (CR) used during the learning of the features. See text for more details.}
\begin{tabular}{lllll}
\toprule
&  \multicolumn {2}{c}{IIIT5K} &  \multicolumn {2}{c}{SVT}\\
\cmidrule(r){2-3} \cmidrule(r){4-5}
& mAP & P@1 & mAP & P@1\\
SIFT + FV & 25.52 & 46.34 & 23.20 & 30.82 \\
Unsup. mid-level + FV & 20.32 & 38.10 & 18.38 & 25.68\\
Sup. mid-level ($\text{CR} = 1 \times 1$) +  FV & 33.96 & 52.20 & 29.75 & 37.46 \\
Sup. mid-level ($\text{CR} = 2\times 2$) + FV &  41.35 & 60.20 & 34.29 & 42.60\\
Sup. mid-level ($\text{CR} = 3\times 3$) + FV & 42.73 & 61.48 & 37.38 & 45.92\\
Sup. mid-level ($\text{CR} = 4\times 4$) + FV & \textbf{43.34} & \textbf{61.72} & \textbf{37.98} & \textbf{46.83}\\
\bottomrule
\end{tabular}
\label{tab:baseline}
\end{table}

\begin{figure*}[t!]
\centering
\includegraphics[width=0.93\linewidth]{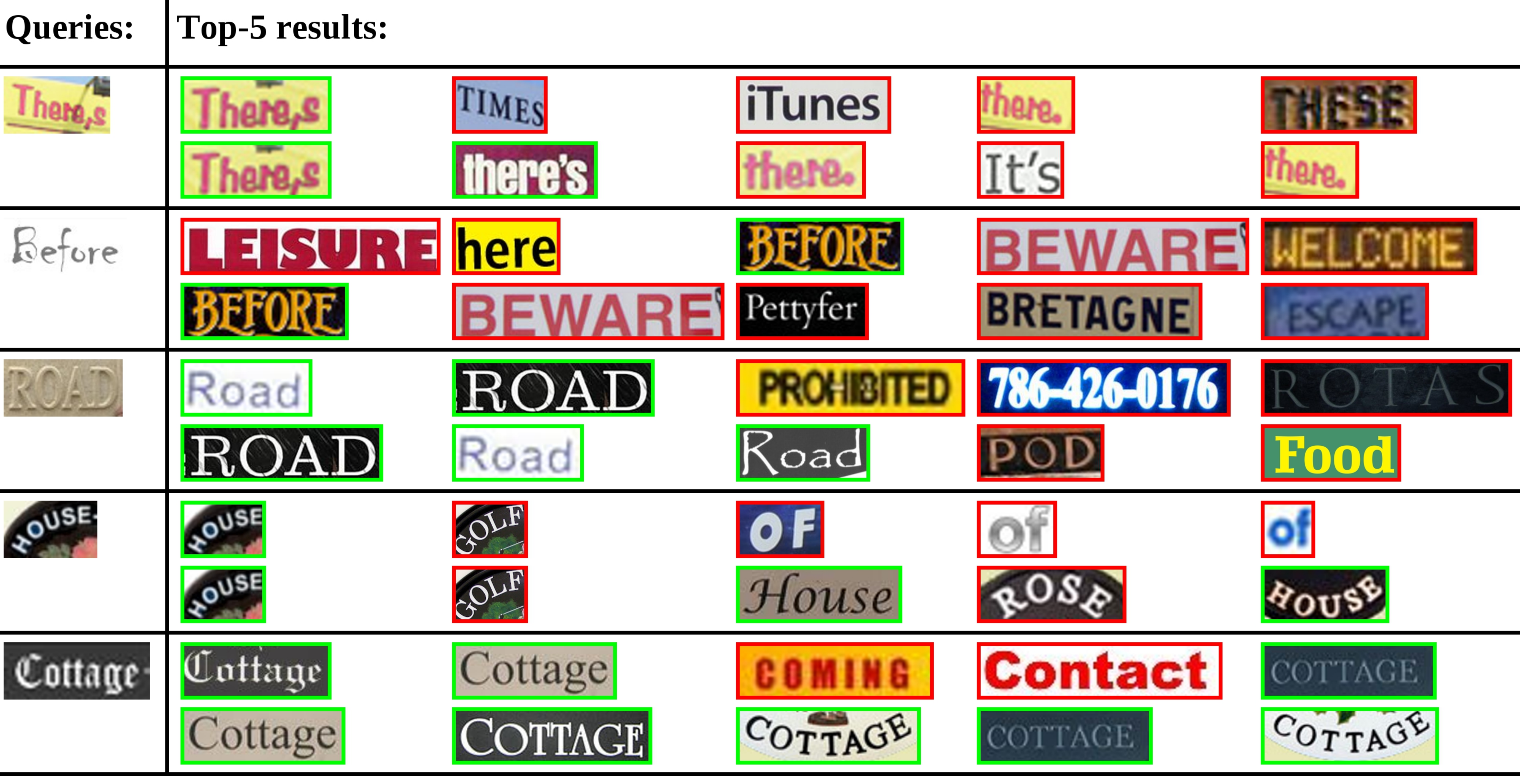}
\caption{Qualitative query-by-example results for some IIIT5K queries. For each query, the first row displays the top retrieved images using SIFT word attributes, and the second row displays the top retrieved images using the proposed mid-level features with word attributes. Correct results are outlined in green.}
\label{fig:qualitative}
\end{figure*}

In the second set of experiments, we are interested in measuring how state-of-the-art global word image representations can benefit from these supervised features. In particular, we focus on the recent word attributes work of Almaz\'an \emph{et al.} \cite{Almazan:2014}, described in Section \ref{sec:representation}.
We also explore the option of combining both FV representations (one based on SIFT and the other based on mid-level features), since their information may be complementary. To do so, we concatenate the global representations based on SIFT and mid-level features before learning the word attributes.
To ensure the fairness of the comparisons, we learned the word attributes of \cite{Almazan:2014} using a dataset comprised of the learning dataset described in Section \ref{sec:datasets} plus IIIT5K and SVT (excluding the test set of the target dataset). In this manner, word attributes based on SIFT and word attributes based on mid-level features have been trained using exactly the same images, and only the type of annotations (text transcriptions only vs text transcriptions and character bounding boxes) differs.

As in \cite{Almazan:2014}, we evaluate on three tasks: query-by-example (QBE) -- \ie, image-to-image retrieval --, query-by-string (QBS) -- \ie, text-to-image --, and recognition --\ie, image-to-text. 
The accuracy of the QBE and QBS tasks is measured in terms of mean average precision.
As  is standard practice, we evaluate recognition using the small lexicon (SL) in IIIT5K and SVT, and the medium lexicon (ML) in IIIT5K. We also evaluate on our more challenging combined lexicon (CL).
The recognition task is measured in terms of precision at $1$.

\begin{table}[t!]
\centering
\scriptsize
\caption{Retrieval results on IIIT5K and SVT datasets on the query-by-example (QBE) and query-by-string (QBS) tasks.}
\begin{tabular}{l|l|cc}
\toprule
Dataset& Method & QBE & QBS  \\
\midrule
\multirow{4}{*}{IIIT5K} & Label Embedding \cite{Rodriguez:2013} & 43.70 & - \\
& [SIFT] + FV + Atts \cite{Almazan:2014} & 68.37 & 72.62\\
& [\textbf{Prop. Mid-features}] + FV +  Atts  & 75.77 & 78.61\\
& [\textbf{Prop. Mid-features} + SIFT] + FV +  Atts  & \textbf{75.91} & \textbf{78.62}\\
\midrule
\multirow{3}{*}{SVT} & [SIFT] + FV + Atts \cite{Almazan:2014} & 60.20 & 81.96\\
& [\textbf{Prop. Mid-features}] + FV + Atts  & 62.68 & 83.83 \\
& [\textbf{Prop. Mid-features} + SIFT] + FV +  Atts  & \textbf{65.94} & \textbf{85.36} \\
\bottomrule
\end{tabular}
\label{tab:retr}
\end{table}

\textbf{Retrieval results.} In Table \ref{tab:retr} we report the retrieval results on IIIT5K and SVT.
On both datasets using supervised mid-level features significantly improves over using SIFT features directly both on the query-by-example and query-by-string tasks.
Combining the mid-level features and SIFT yields even further improvements on SVT. To the best of our knowledge, the best reported results on query-by-example and query-by-string on both datasets were those of \cite{Almazan:2014}, and using mid-level features significantly improves those results.

\textbf{Recognition results.} Table \ref{tab:rec} shows the recognition results of our approach compared to the state-of-the-art. 
On IIIT5K, \cite{Almazan:2014} held the best results, and these results are improved thanks to the mid-level features.
In the more difficult medium and combined lexicons, these improvements are very noticeable: from $82.07$ to $85.93$ and from $77.77$ to $83.03$.
A similar trend can be observed on SVT, where the improvement on the combined lexicon is very significant.
Increasing the output dimensionality from $96$ to $192$ dimensions also significantly improves the accuracy on the SVT small lexicon recognition task up to an $91.81\%$. We only observed this behavior in this particular case; other datasets and tasks do not require larger representations. 

\begin{figure}
\centering
\includegraphics[width=0.49\linewidth,trim=0.45cm 0cm 0.15cm 0.2cm,clip=true]{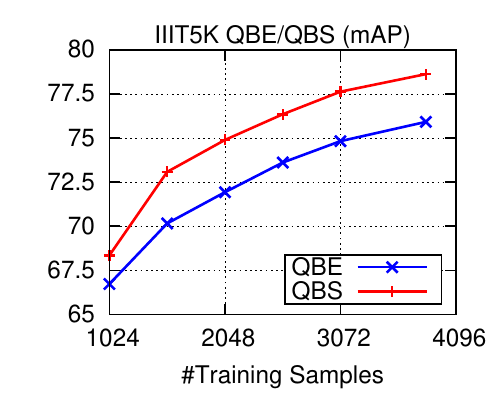}
\includegraphics[width=0.49\linewidth,trim=0.45cm 0cm 0.15cm 0.2cm,clip=true]{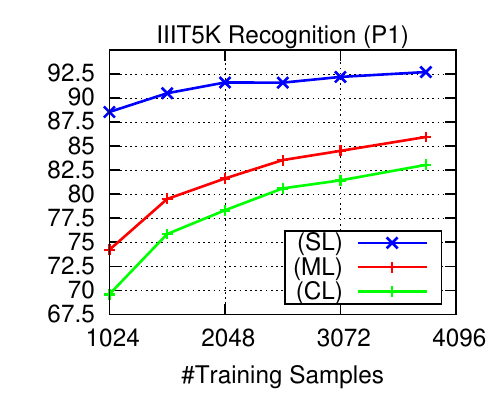}
\caption{Accuracy on the IIIT5K dataset as a function of the number of annotated training words used to learn the mid-level space and the word attributes.
Left: retrieval accuracy for the query-by-example (QBE) and query-by-string (QBS) tasks. Right: recognition accuracy using a small-sized lexicon (SL), a medium-sized lexicon (ML), and a combined lexicon (CL).
In all cases the accuracy has not yet saturated, and more training data will likely improve the results.}
\label{fig:limitedtraining}
\end{figure}

The best reported results on SVT are those of \cite{Bissacco:2013} and \cite{Jaderberg:2014b}, both of which use millions of training samples. Using a fraction of the training data we obtain results better than PhotoOCR \cite{Bissacco:2013}, but are outperformed by Jaderberg \emph{et al.} \cite{Jaderberg:2014b}. However, as seen on Figure \ref{fig:limitedtraining}, the performance of our method has not saturated: using more data as \cite{Jaderberg:2014b} does will likely increase the accuracy on both datasets.
Our method also has other advantages such as leading to tiny ($96$-$192$ dimensions) signatures that can be used for image-to-image and text-to-image matching.
Compared to methods that do not use millions of training samples \cite{Mishra:2012b, Goel:2013, Yao:2014, Jaderberg:2014a} , our results are significantly better.

\begin{table}[t!]
\centering
\scriptsize
\caption{Comparison with the state-of-the-art on recognition accuracy on the IIIT5K and SVT datasets with small (SL), medium (ML), and combined (CL) lexicons. Methods marked with an $*$ use several millions of training samples.}
\begin{tabular}{p{0.65cm}|l|p{0.35cm} p{0.35cm} p{0.35cm}}
\toprule
Dataset& Method & SL & ML & CL \\
\midrule
\multirow{6}{*}{IIIT5K} & High Order Language Priors \cite{Mishra:2012a} & 64.10 & 57.50 & - \\
& Label Embedding \cite{Rodriguez:2013} & 76.10 & 57.40 & -\\
& Strokelets \cite{Yao:2014} & 80.20 & 69.3 & -\\
& [SIFT] + FV +  Atts \cite{Almazan:2014} & 91.20 & 82.07 & 77.77\\
& [\textbf{Prop. Mid-features}] + FV + Atts  & 92.67 & 85.93 & 83.03\\
& [\textbf{Prop. Mid-features} + SIFT] + FV + Atts  & \textbf{93.27} & \textbf{86.57} & \textbf{83.07}\\

\midrule
\multirow{10}{*}{SVT} & ABBY \cite{Goel:2013} & 35.00 & - & -\\
& Mishra \etal \cite{Mishra:2012b} & 73.26 &- & -\\
& Synthesized Queries \cite{Goel:2013} & 77.28 &- & -\\
& Strokelets  \cite{Yao:2014} & 75.89 &- & -\\
& *PhotoOCR \cite{Bissacco:2013} \emph{(in house training data)} & 90.39 & - & -\\
& Deep CNN \cite{Jaderberg:2014a} &   86.1 & - & -\\
& *Deep CNN \cite{Jaderberg:2014b} \emph{(synthetic training data)}& \textbf{95.4} & - & -\\
& [SIFT] + FV + Atts \cite{Almazan:2014} & 89.18 &- & 72.49\\
& [\textbf{Prop. Mid-features}] + FV + Atts  & 89.49 & - & 73.42 \\
& [\textbf{Prop. Mid-features} + SIFT] + FV + Atts  & 90.73 & - & \textbf{76.51} \\
& [\textbf{Prop. Mid-features} + SIFT] + FV + Atts (192d)  & 91.81 & - & \textbf{76.51} \\

\bottomrule
\end{tabular}
\label{tab:rec}
\end{table}

\textbf{Qualitative results.} Figure \ref{fig:qualitative} shows qualitative results of the query-by-example task for some IIIT5K queries. 
In the more difficult queries (\eg ``Before'' or ``HOUSE'') the mid-level features are clearly superior. In general, even when the retrieved results are not correct, they are closer to the query than when using SIFT features directly.
\section{Conclusions}
\label{sec:conclusions}
In this paper we have introduced supervised mid-level features for the task of word image representation.
These features are learned by leveraging character bounding box annotations at training time, and correlate visual blocks with the characters (and, most importantly, the character regions) in which such blocks tend to appear.
Despite using character information at training time, one key advantage of our approach is that it does not require localizing characters explicitly at testing time.
Instead, our mid-level features can be densely extracted in an efficient manner.
We used these mid-level features as a building block of the word attributes framework of \cite{Almazan:2014}.
The proposed mid-level features outperform equivalent representations based on SIFT on two standard benchmarks using tiny signatures of only $96$ dimensions, and obtain state-of-the-art results on retrieval and recognition tasks.
We finally note that the proposed approach can be seen as a way to learn mid-level features from annotated parts at training time (characters and character regions in our case) without explicitly localizing them at testing time.
We believe that the key ideas behind these mid-level features are not limited to text, and could be exploited well beyond the scene-text domain, \eg, for generic object categorization.

\appendix
\section{Efficient Word Representation}
\label{app:efficient}
This appendix describes an approach to compute all mid-level features of a new image exactly in an efficient manner.

Once the matrix $\mathbf{U}$ has been learned as described in Section \ref{sec:learning}, one can use it to compute the mid-level features of a new image. A naive process would be as follows: i) Using a given stepsize (\eg $4$ pixels), extract all possible blocks at all sizes (see parameters used during the offline learning step). This leads to up to tens of thousands of overlapping blocks per image. ii) For each block, extract SIFT descriptors, compute a FV with a $2 \times 2$ spatial pyramid, and $\ell_2$ normalize. iii) Project the FVs with $\mathbf{U}$, and $\ell_2$  normalize again -- $\ell_2$  normalization after PCA/CCA usually improves the results since it accounts for the missing energy in the dimensionality reduction step \cite{Jegou:2012}. This was shown in Algorithm \ref{alg:alg2}.

Unfortunately, this process is not feasible in practice, as it would take a long time to compute and project all the FVs for all the blocks independently in a naive way.
Instead, we propose an approach to compute these descriptors exactly in an efficient manner.
The key idea is to isolate the contribution of each individual SIFT feature towards the mid-level feature of a block that included such a SIFT feature.
If that contribution is linear, one can compute the individual contribution of each SIFT feature in the image and accumulate them in an integral representation. Then, the mid-level representation of an arbitrary block could be computed with just $2$ additions and $2$ subtractions over vectors, and  computing all possible descriptors of all possible blocks becomes feasible and efficient.

At first sight the contribution of the features is not linear due to two reasons: the spatial pyramid, and the $\ell_2$  normalization of the FVs \emph{before} the projection with $\mathbf{U}$. Fortunately, both problems can be overcome.
Our approach works thanks to three key properties:

\textbf{1) The FV is additive} when not performing any normalization. Given a set of descriptors $\mathcal{S}$, $fv(\mathcal{S}) = \frac{1}{|\mathcal{S}|}\sum_{s \in \mathcal{S}} fv(s)$, since the FV aggregates the encoded descriptors using average pooling. This is key since it implies that we can compute FVs independently and then aggregate them. This is also true if the descriptors are projected linearly, \ie $\mathbf{U}^T fv(\mathcal{S}) = \frac{1}{|\mathcal{S}|} \sum_{s \in \mathcal{S}} \mathbf{U}^T fv(s)$. However, it is \emph{not} true if we $\ell_2$  normalize the FVs, \ie, $fv(\mathcal{S})/||fv(\mathcal{S})||_2 \neq  \frac{1}{|\mathcal{S}|}   \sum_{s \in \mathcal{S}} fv(s)/||fv(s)||_2$. Fortunately, $\ell_2$  normalizing the FVs will not be necessary, \cf next property.

\textbf{2) The normalization of the FV is absorbed by the normalization after the projection with $\mathbf{U}$}. If we let $f$ be an unnormalized FV and let $f_2$ be the $\ell_2$  normalized version, then $\mathbf{U}^T f_2/||\mathbf{U}^T f_2||_2 = \mathbf{U}^T f/||\mathbf{U}^T f||_2$. Even if we $\ell_2$  normalize the FVs when learning the projection with CCA (since $\ell_2$  normalization greatly increases its discriminative power), it is not necessary to $\ell_2$  normalize them when representing the words if they are also going to be normalized after projection, which makes the aggregation of FVs seen in property $1$ possible. Note how the $\frac{1}{|\mathcal{S}|}$ factor is also absorved by the $\ell_2$ normalization.

\textbf{3) The projection of a FV with spatial pyramid is also additive.} Projecting a FV $f$ with spatial pyramid with $\mathbf{U}$ is equivalent to projecting each spatial region independently with the corresponding rows of $\mathbf{U}$ and then aggregating the results. Assuming spatial pyramids of $2\times 2$, it is also possible to rearrange $\mathbf{U}$ into $\hat{\mathbf{U}} \in \Re^{\frac{D_v}{4} \times 4K}$. In this case, projecting a FV \emph{without spatial pyramid} with $\hat{\mathbf{U}}$ leads to a descriptor of $4K$ dimensions. Each group of $K$ dimensions represents the results of projecting $f$ assuming that it was representing one of the 4 spatial quadrants of a larger block.

\begin{figure}
\centering
\includegraphics[width=0.65\linewidth]{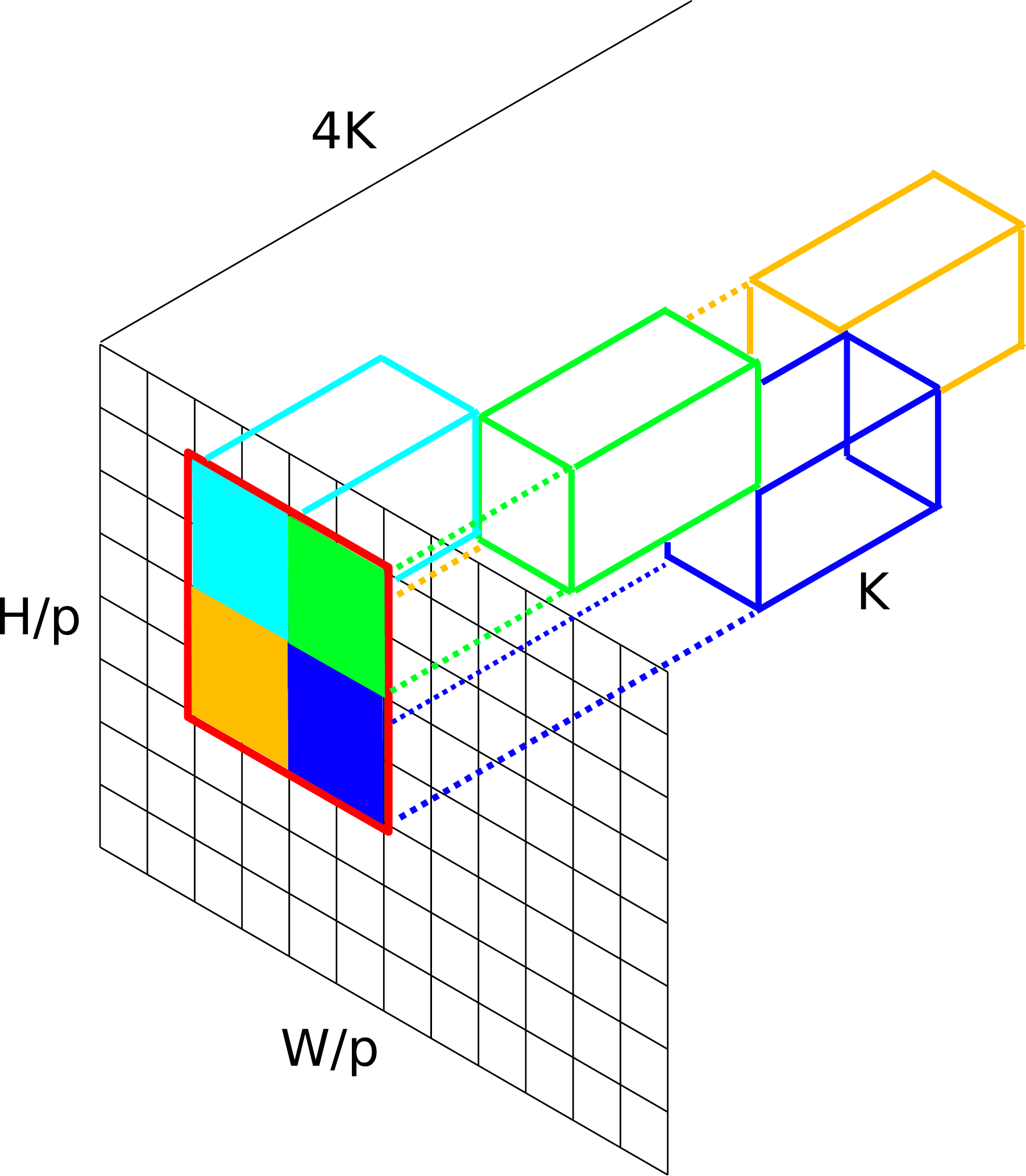}
\caption{Grid $\mathcal{G}$ representing the influence of each descriptor in an image towards the mid-level features. To compute the mid-level descriptor of a block  (\eg the red block), one has to sum the $K$-dimensional descriptors inside the colored cuboids --done efficiently using an integral representation--, and $\ell_2$ normalize the final result.}
\label{fig:depth}
\end{figure}

By exploiting these properties, we can compute the mid-level descriptors of all possible blocks in an image in an efficient manner.
In our approach, we begin by dividing the target image in contiguous, non-overlapping cells of $p\times p$ pixels, where $p$ controls the step size between two consecutive blocks. In our experiments we set $p=4$. We compute a FV in each of these regions \emph{with neither spatial pyramids nor $\ell_2$ normalization} and project them with $\hat{\mathbf{U}}$ into a space of $4K$ dimensions. This leads to a grid representation of the image $\mathcal{G} \in \Re^{\frac{H}{p} \times \frac{W}{p} \times 4K}$, where $H$ and $W$ are the height and the width of the image. The ``depth'' of the representation can be separated into $4$ groups that represent the projection of that particular grid cell depending on its position on the pyramid. Figure \ref{fig:depth} illustrates this. 
Note that, since the cells do not overlap, computing the FVs of all those cells has approximately the same cost as computing one single FV using the descriptors of all the image, and does not involve any particular extra cost. Keeping in memory $\frac{H}{p} \times \frac{W}{p} \times 4K$ elements at the same time is also not an issue.
Finally, we compute an integral representation over the height and the width, \ie, $\hat{\mathcal{G}}_{i,j,k} = \sum_{1\leq a \leq i, 1\leq b \leq j} \mathcal{G}_{a,b,k}$.

At this point, one can easily compute the descriptor of any given block by i) separating the block in $4$ different spatial regions. ii) Computing the descriptor of each region independently with two sums and two subtractions on vectors of dimension $K$ by accessing the corresponding rows and columns of the integral representation $\hat{\mathcal{G}}$, and by keeping only the group of $K$ dimensions corresponding to the particular spatial regions. iii) Aggregating the  descriptors of each spatial region, and iv) $\ell_2$ normalizing the final result.

In our experiments we use images of $120$ pixels in height, a step size of $p=4$, a block spatial pyramid of $2\times 2$, and $K=62$ projections.
We also extract blocks at $5$ different block sizes: $16\times 16$, $24 \times 24$, \ldots, $48 \times 48$.
With this setup, we can extract and describe all blocks in an image in less than a second using a MATLAB implementation and a single core.

{\small
\bibliographystyle{ieee}
\bibliography{egbib}
}

\end{document}